\documentclass[letterpaper]{article} % DO NOT CHANGE THIS
\usepackage{aaai24}  % DO NOT CHANGE THIS
\usepackage{times}  % DO NOT CHANGE THIS
\usepackage{helvet}  % DO NOT CHANGE THIS
\usepackage{courier}  % DO NOT CHANGE THIS
\usepackage[hyphens]{url}  % DO NOT CHANGE THIS
\usepackage{graphicx} % DO NOT CHANGE THIS
\usepackage{natbib}  % DO NOT CHANGE THIS AND DO NOT ADD ANY OPTIONS TO IT
\usepackage{caption} % DO NOT CHANGE THIS AND DO NOT ADD ANY OPTIONS TO IT
\usepackage{algorithm}
\usepackage{algorithmic}
\usepackage{amsmath,amssymb,amsfonts,amsthm}
\usepackage{bm}
\usepackage{booktabs}
\usepackage{multirow}
\usepackage{subfigure}
\usepackage{color}

\theoremstyle{definition}

\newcommand{\ourmodel}{\text{DuSkill}}

\newcommand{\Real}{\mathbb{R}}

\newcommand{\StateSet}{S}
\newcommand{\ActionSet}{A}
\newcommand{\Dynamics}{P}
\newcommand{\RewardFunc}{r}
\DeclareMathOperator*{\argmax}{argmax}

\newcommand{\Proto}{\mathcal{Z}_\rho}
\newcommand{\Skill}{\mathcal{Z}_\sigma}
\newcommand{\proto}{z_\rho}
\newcommand{\skill}{z_\sigma}
\newcommand{\protoenc}{q_{\rho}}
\newcommand{\skillenc}{q_{\sigma}}
\newcommand{\protopri}{p_{\rho}}
\newcommand{\skillpri}{p_{\sigma}}
\newcommand{\protodec}{\epsilon_{\rho}}
\newcommand{\skilldec}{\epsilon_{\sigma}}

\usepackage{newfloat}
\usepackage{listings}
\DeclareCaptionStyle{ruled}{labelfont=normalfont,labelsep=colon,strut=off} % DO NOT CHANGE THIS
\floatstyle{ruled}
\newfloat{listing}{tb}{lst}{}
\floatname{listing}{Listing}
\title{Robust Policy Learning via Offline Skill Diffusion}
\author{
    Woo Kyung Kim,
    Minjong Yoo,
    Honguk Woo\thanks{Honguk Woo is the corresponding author.}
}
\affiliations{
    Department of Computer Science and Engineering, Sungkyunkwan University\\
    \{kwk2696, mjyoo2, hwoo\}@skku.edu
}

\usepackage{bibentry}
\begin{document}

\maketitle

\begin{abstract}
Skill-based reinforcement learning (RL) approaches have shown considerable promise, especially in solving long-horizon tasks via hierarchical structures. 
These skills, learned task-agnostically from offline datasets, can accelerate the policy learning process for new tasks. Yet, the application of these skills in different domains remains restricted due to their inherent dependency on the datasets, which poses a challenge when attempting to learn a skill-based policy via RL for a target domain different from the datasets' domains. 
In this paper, we present a novel offline skill learning framework $\ourmodel$ which employs a guided Diffusion model to generate versatile skills extended from the limited skills in datasets, thereby enhancing the robustness of policy learning for tasks in different domains. 
Specifically, we devise a guided diffusion-based skill decoder in conjunction with the hierarchical encoding to disentangle the skill embedding space into two distinct representations, one for encapsulating domain-invariant behaviors and the other for delineating the factors that induce domain variations in the behaviors. 
Our $\ourmodel$ framework enhances the diversity of skills learned offline, thus enabling to accelerate the learning procedure of high-level policies for different domains.
Through experiments, we show that $\ourmodel$ outperforms other skill-based imitation learning and RL algorithms for several long-horizon tasks, demonstrating its benefits in few-shot imitation and online RL.
\end{abstract}

% % % % %
\section{Introduction}
Skill-based learning demonstrates the potentials in accelerating the adaptation to complex long-horizon tasks by leveraging pretrained skill representations on behavior patterns from the offline datasets. However, existing approaches in skill-based reinforcement learning (RL) (e.g.,~\citealt{sprl:spirl,sprl:skild}) and skill-based few-shot imitation learning (e.g.,~\citealt{sprl:fist,sprl:ret}) often operate under the premise that the target domain for a downstream task was present during skill pretraining. Thus, policies learned with the pretrained skills might lead to sub-optimal performance, particularly when the target domain diverges from the domains of the given datasets. 

% Figure 0
\begin{figure}[ht] 
    \centering
        \includegraphics[width=.98\columnwidth]{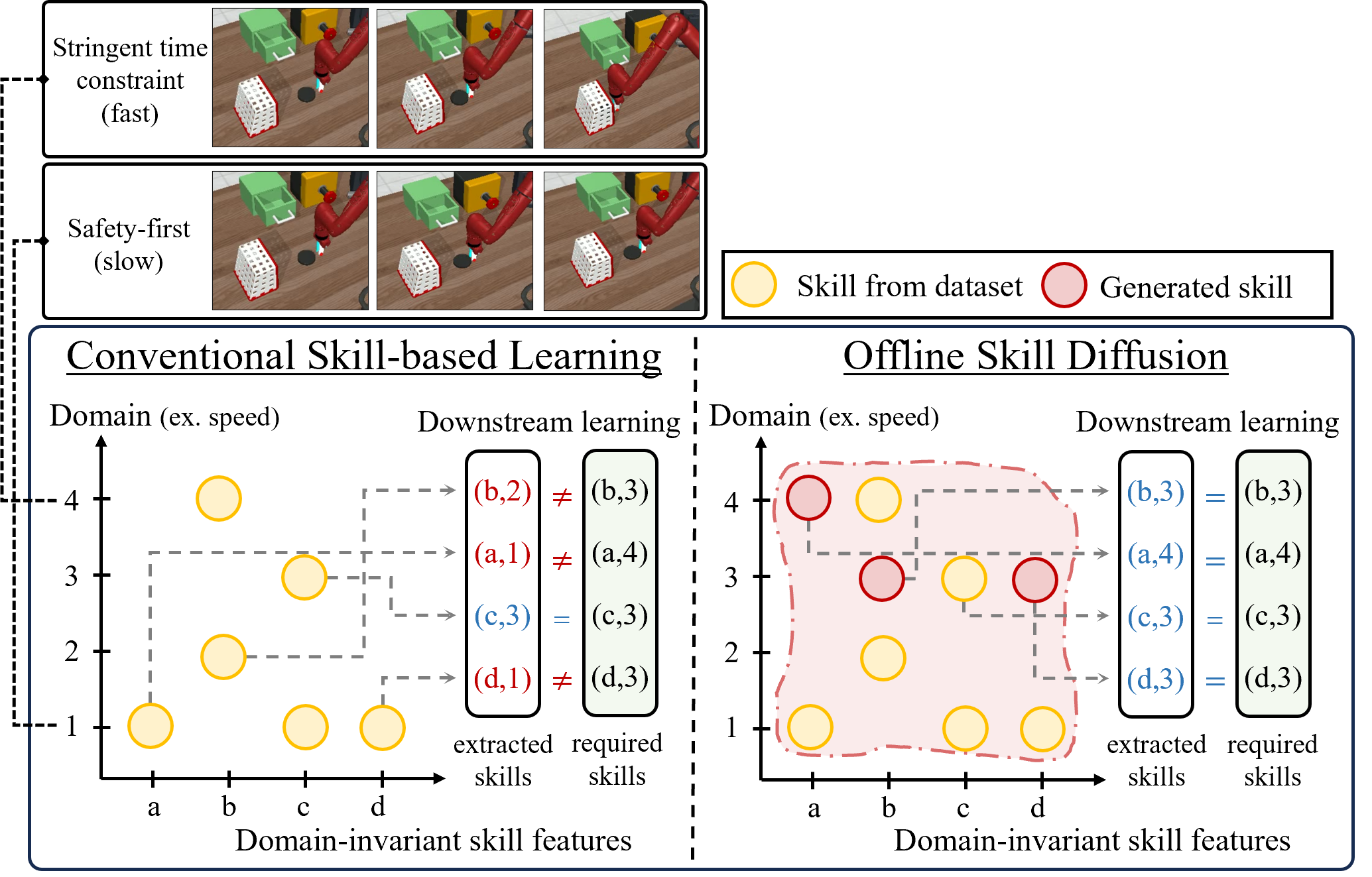}
    \caption{Concept of Offline Skill Diffusion: When a downstream task belongs to the domain different from those of the training datasets, conventional skill-based learning approaches struggle in learning and choosing suitable skills. In contrast, our offline skill diffusion expands the skill diversity that goes beyond the training datasets, enabling the execution of compatible skills for the downstream task. The skills are discretely represented for visual illustration.}
    \label{fig:problem}
\end{figure}

As shown in the left side of Figure~\ref{fig:problem}, where each small circle denotes a specific skill, conventional skill learning approaches might experience low performance in the high-level policy for a downstream task, if the task calls for skills that differ from the pretrained ones. For instance, suppose that robotic manipulation skills are learned from the datasets in the safety-first domain; then, they might heavily lean towards ``slow'' speed manipulation. In that case, a high-level policy learned with these skills might fail to adapt efficiently to the downstream task that involves stringent time constraints. These situations often arise in the environment encompassing diverse domains, as a single task can require different skills depending on the domain it is in. The dependency of conventional skill-based learning approaches on the specificity of the datasets exacerbates these challenges. Moreover, it is practically difficult to obtain comprehensive datasets that span all potential skills for diverse domains. 

To tackle these challenges in skill-based learning, we take a novel approach, offline skill diffusion, aiming to broaden skill diversity beyond mere imitation from the datasets. Given that diffusion models have been recognized for their efficacy in generating human-like images with conditional values~\cite{du:stabledu,du:cfdg,du:autodu}, we leverage a guided diffusion model for the skill decoder to generate diverse skills. The right side of Figure~\ref{fig:problem} illustrates the benefits of our framework approach, where the red-colored dotted quadrangle represents an expanded skill set encompassing all the skills that are required to solve the downstream task in a different domain. 

In this paper, we present the $\ourmodel$ framework, designed to generate diverse skills for downstream tasks that can belong to the domains different from the source domains in the training datasets. Recognizing that certain aspects of a skill remain consistent despite domain variations, we view each skill as a composition of domain-invariant and domain-variant features. Then, we employ a guided diffusion-based decoder along with a hierarchical domain encoder so as to effectively disentangle each skill into two separate embedding spaces.  
The hierarchical domain encoder systematically segments skills into domain-invariant and domain-variant embeddings by conditioning only the lower encoder on domain variations. With two distinct embeddings, the conditional generation process of the guided diffusion-based decoder enables distinct modulation of each embedding, thereby facilitating the execution of a wide range of skills in different domains.

For downstream tasks in different domains, we train a high-level policy which produces both domain-variant and domain-invariant embeddings. The high-level policy operates alongside the frozen guided diffusion-based decoder, which encompasses the necessary knowledge for generating a broad range of skills, adaptable to various domains. As such, our proposed framework stands apart from existing skill-based learning approaches~\cite{sprl:spirl,sprl:skild}, as it enables the generation of diverse skills that extend beyond the training datasets.

The contributions of our work are summarized as follows.
\begin{itemize}
    \item We present the $\ourmodel$ framework, which facilitates robust skill-based policy learning for downstream tasks in different domains. 
    \item We develop the offline skill diffusion model, incorporating the hierarchical domain encoder and guided diffusion-based decoder. The model enables the diverse skill generation that extends beyond the training datasets.
    \item We test the framework with long-horizon tasks in various domains, demonstrating its capability to adapt to different domains in both few-shot imitation and online RL settings.
\end{itemize}

% % % % %
\section{Preliminaries and Problem Formulation}
Given the training datasets $\mathcal{D}=\{\tau_i\}_{i \leq n}$, where each trajectory $\tau_i$ in the datasets is represented as a sequence of state and action pairs $\{(s_t,a_t)\}_{t \leq H}$, the objective of skill representation learning is to accelerate the adaptation to long-horizon tasks by leveraging pretrained skill representations.
Here, a skill is defined as a sequence of $h$ consecutive actions $\bm{a}=\{a_{t},...,a_{t+h}\}$~\cite{sprl:spirl,sprl:fist}.
Through the joint learning of both a skill encoder $q(z|\bm{a})$ and a skill decoder $\epsilon(\bm{a}|z)$ on the datasets, we are able to obtain a skill embedding $z \in \mathcal{Z}$.
The learning objective involves optimizing the evidence lower bound (ELBO), which consists of a reconstruction term and a regularization term,
\begin{equation}\label{equ:obj:vae}
    \mathcal{L}_{\text{VAE}} = \mathbb{E}_{\bm{a} \sim \mathcal{D}} \left[ -\log \epsilon(\bm{a}|z) + \beta D_{\text{KL}}(p(z)||q(z|\bm{a}) \right]
\end{equation}
where $D_{\text{KL}}$ is the Kullback-Leibler (KL) divergence, $p(z)$ is a prior following a unit Gaussian distribution $\mathcal{N}(0,I)$ and $\beta$ is a hyperparameter for regularization~\cite{vae:bvae}.
Then, this skill representation is leveraged to accelerate the downstream task adaptation.

\noindent \textbf{Problem Formulation.} In our problem formulation, we operate under the premise that training datasets $\mathcal{D}$ are available.
As skills required for a task might vary depending on the domain it belongs to, we further assume that the datasets are collected from multiple source domains.
The variations across the domains are parameterized by $\omega \in \Omega$, and this parameter information is incorporated in the datasets.
Then, we aim to tackle downstream tasks in a range of domains distinct from the source domains. Successfully achieving this requires more than merely imitating the skills present in the given datasets, due to the varied nature of the tasks across different domains. 

We formulate a downstream task as a goal-conditioned Markov decision process (MDP) $\mathcal{M}$ combined with its domain $\Omega$.
The goal-conditioned MDP $\mathcal{M}$ is denoted as $(\StateSet, \ActionSet, \Dynamics, \RewardFunc, \mathcal{G}, \gamma, \mu_0)$ where $s \in \StateSet$ is a state space, $a \in \ActionSet$ is an action space, $\mathcal{G}$ is a goal space, $\Dynamics: \StateSet \times \ActionSet \times \Omega \rightarrow [0,1]$ is a transition probability, $\RewardFunc: \StateSet \times \ActionSet \times \mathcal{G} \times \Omega \rightarrow \Real$ is a reward function, $\gamma \in [0, 1]$ is a discount factor, and $\mu_0: \StateSet \rightarrow [0, 1]$ is an initial state distribution. 
The domain variations may affect either the reward function or the transition probability in an MDP, while the goal space remains consistent. 
For instance, on the top of Figure~\ref{fig:problem}, a task may have the same goal space as slide puck in the goalposts, but the domain related to the time constraint results in different reward functions.

Then, our objective is to maximize the discounted cumulative sum of rewards for downstream tasks in different domains, through a high-level policy $\pi(z|s)$, 
\begin{equation}
\pi^* = \argmax_\pi \mathbb{E}_{z\sim\pi(\cdot|s), \bm{a}\sim\epsilon(\cdot|z)} \left[ \sum_{t=0}^{T-1} \gamma^t r\left(s_t, a_t)\right) \right]
\end{equation}
where $\epsilon(\bm{a}|z)$ is a skill decoder and $T$ is the maximum length for an episode.

% Figure 1
\begin{figure}[ht] 
    \centering
        \includegraphics[width=0.98\columnwidth]{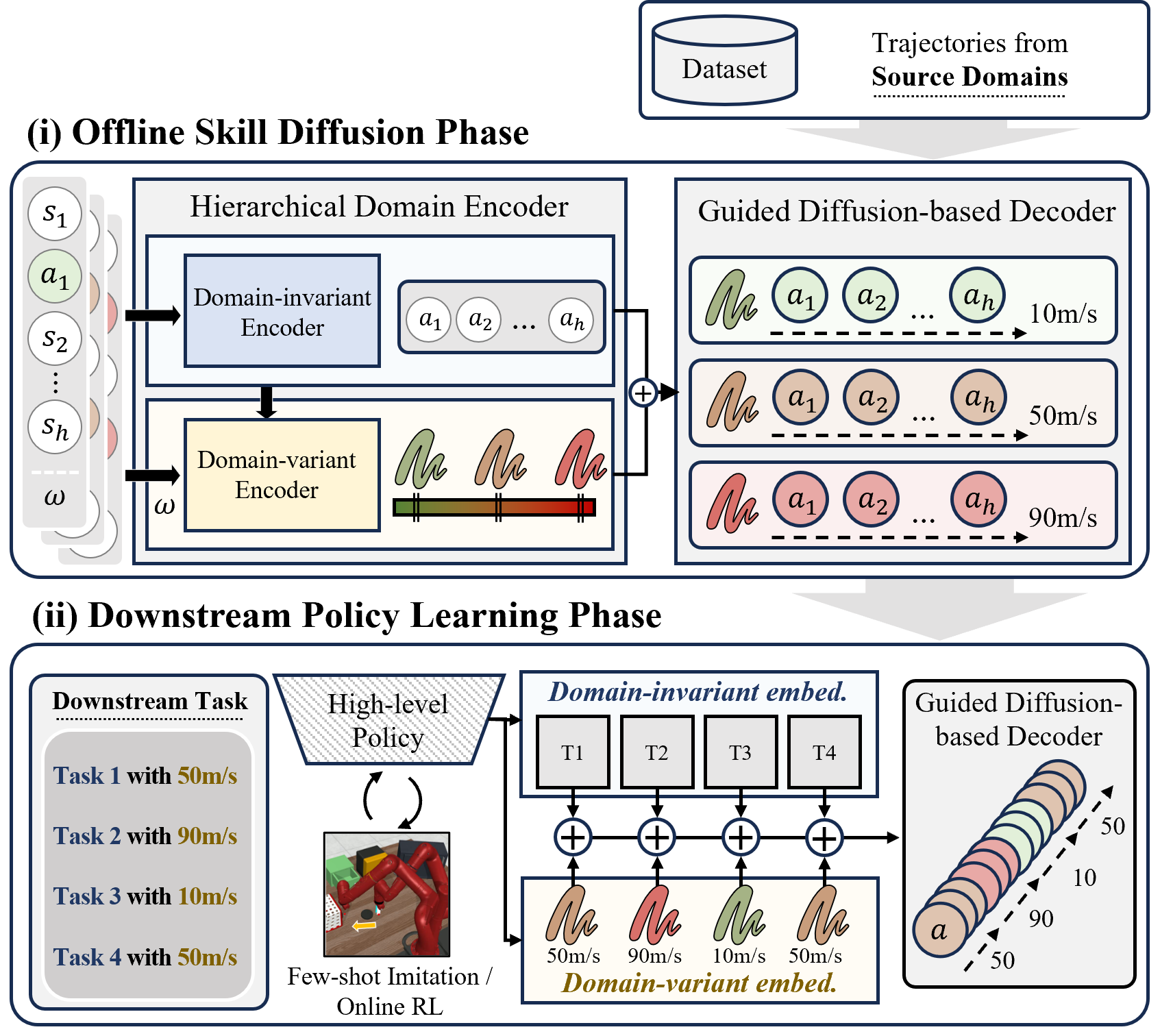}
    \caption{Offline Skill Diffusion and Downstream Policy Learning in $\ourmodel$: (\romannumeral 1) In the offline skill diffusion phase, a skill is decomposed into domain-invariant and domain-variant embeddings, and then they are combined through the guided diffusion based decoder to generate diverse skills. (\romannumeral 2) In the downstream policy learning phase, a high-level policy is learned for a task in different domains either through few-shot imitation or online RL.}
    \label{fig:overall}
\end{figure}
% % % % %
\section{Approach}
\subsection{$\ourmodel$ Framework}
% Figure 2
\begin{figure*}[ht] 
    \centering
        \includegraphics[width=.98\textwidth]{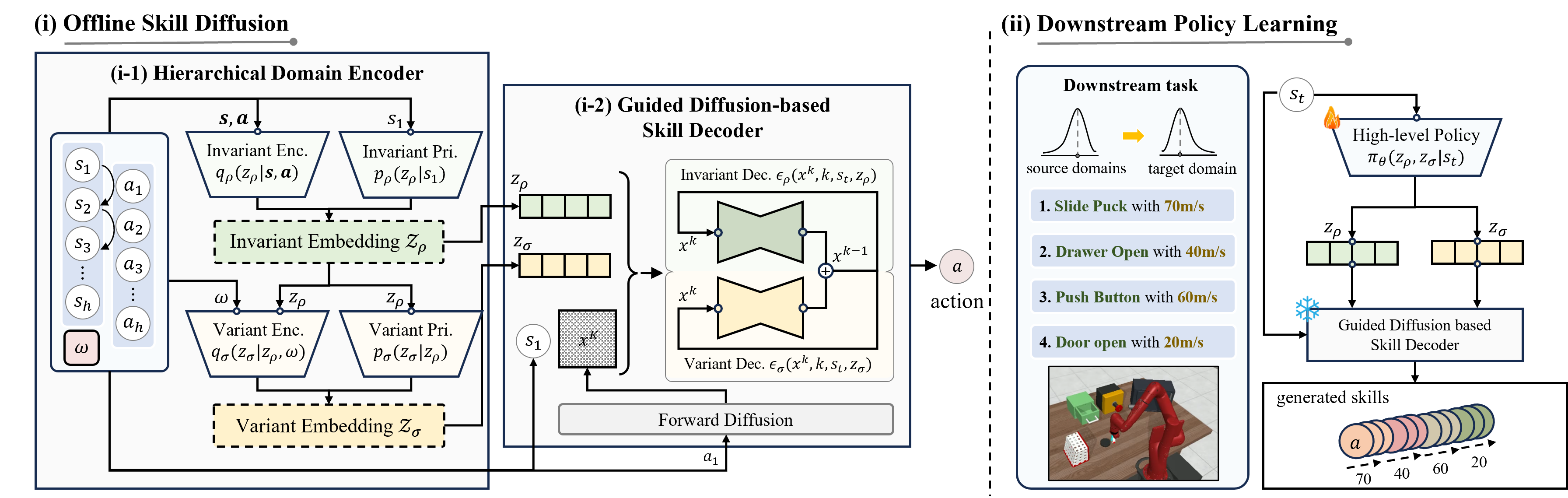}
    \caption{$\ourmodel$ Framework: In (\romannumeral 1-1), the hierarchical domain encoder disentangles the domain-invariant and domain-variant embeddings. At the same time, the domain-invariant and domain-variant priors are jointly learned with these encoders. For diverse skill generation, the encoders are trained in conjunction with the guided diffusion-based decoder in (\romannumeral 1-2). Here, the domain-invariant decoder and domain-variant decoder are responsible for reconstructing actions based on the domain-invariant and domain-variant embeddings, respectively. In (\romannumeral 2), the high-level policy is learned to solve the task in different domain either through few-shot imitation or online RL.}
    \label{fig:framework}
\end{figure*}
To facilitate the diverse skill generation, we propose the $\ourmodel$ framework consisting of two main phases: (\romannumeral 1) the offline skill diffusion phase, and 
%
% 0815 adaptation 이 아니라, policy learning 인듯. 
(\romannumeral 2) the downstream policy learning phase, as illustrated in Figure~\ref{fig:overall}.

In the offline skill diffusion phase, we employ a guided diffusion-based decoder in conjunction with a hierarchical domain encoder to disentangle skills into domain-invariant and domain-variant embeddings.
Specifically, the hierarchical domain encoder consists of a domain-invariant encoder and a domain-variant encoder.
The domain-invariant encoder processes a sequence of state and action, resulting in domain-invariant embedding.
Meanwhile, the domain-variant encoder takes the domain-invariant embedding and the domain parameterization as input to produce the domain-variant embedding.
In this way, these encoders play distinct roles, in which domain-invariant encoder is responsible for encapsulating features necessary to reconstruct fundamental action sequences pertinent to achieving goals, while the domain-variant encoder is tailored to capture the features related to domain variations. 
To effectively disentangle skill features via the hierarchical domain encoder, we utilize a guided diffusion-based decoder with two components, where one is conditioned on the domain-invariant embedding, while the other focuses on the domain-variant embedding.
This conditional generation mechanism facilitates a distinct influence of the domain-invariant and domain-variant embeddings on the separate segments of generated action sequences.
By employing the hierarchical domain encoder and guided diffusion-based decoder, our framework is capable of generating diverse action sequences that encompass different combinations of the domain-invariant and domain-variant features.

In the downstream policy learning phase, we exploit the disentangled embeddings via a high-level policy, which produces both domain-invariant and domain-variant embeddings.
These embeddings are then fed into the frozen guided diffusion-based decoder, which generates executable skills.
In this phase, we consider few-shot imitation and online RL adaptation scenarios, where the high-level policy adapts to the tasks in different domains through either fine-tuning on a limited number of trajectories or online RL interactions.

\subsection{Offline Skill Diffusion}
\noindent\textbf{Hierarchical domain encoder.} 
To disentangle domain-invariant and domain-variant features from skills, we introduce a hierarchical encoding approach. 
This enables learning in two distinct embedding spaces: the domain-invariant embedding space $\Proto$ and the domain-variant embedding space $\Skill$.
Specifically, we employ a domain-invariant encoder $\protoenc$ which maps a sequence of states and actions to the domain-invariant embedding. We also use a domain-variant encoder $\skillenc$ which maps the domain-invariant embedding and the domain parameterization $\omega$ to the domain-variant embedding, i.e., 
\begin{equation}\label{equ:def:enc}
    \proto \sim \protoenc(\bm{s},\bm{a}), \skill \sim \skillenc(\proto,\omega)
\end{equation}
where $\bm{s}=\{s_t,...,s_{t+h}\}$ is a sequence of states, $\bm{a} = \{a_t,...,a_{t+h}\}$ is a sequence of actions, $\proto \in \Proto$ is the domain-invariant embedding, and $\skill \in \Skill$ is the domain-variant embedding.
To optimize the encoders and the skill decoder $\epsilon$, we employ the evidence lower bound (ELBO) loss using~\eqref{equ:obj:vae}, similar to~\cite{sprl:skild,sprl:fist}, i.e., 
\begin{equation} \label{equ:obj:hvae}
\begin{aligned}
    \mathcal{L}_{\text{HVAE}} = & \mathbb{E}_{(\bm{s}, \bm{a}) \sim \mathcal{D}} \bigg[ {- \prod_{t=0}^{h} \log \epsilon(a_t|s_t,\proto,\skill)} & \\
    & + \beta_\rho D_{\text{KL}} (p(\proto) || \protoenc(\proto | \bm{s},\bm{a})) & \\
    & + \beta_\sigma D_{\text{KL}} (p(\skill | \proto) ||  q(\skill | \proto,\bm{s},\bm{a})) \bigg]
\end{aligned}
\end{equation}
where $q(\skill|\proto,\bm{s},\bm{a})$ is a naive modeling for the domain-variant encoder, and $\beta_\rho$ and $\beta_\sigma$ are regularization hyperparameters. The prior $p(z_\sigma)$ and $p(z_\rho|z_\sigma)$ are set to be unit Gaussian.
To disentangle skill features, we modify the regularization term of the domain-variant encoder in~\eqref{equ:obj:hvae} as 
\begin{equation} \label{equ:obj:asp}
    \mathcal{L}_{\text{aspect}} = \mathbb{E}_{(\bm{s}, \bm{a}) \sim \mathcal{D}} \left[ \log \frac{q(\skill | \proto,\bm{s},\bm{a})}{\skillenc (\skill | \proto,\omega)}\right]. 
\end{equation}
This loss term enables the domain-variant encoder to construct the distinct embedding space $\Skill$, capturing domain-variant features conditioned on the domain-invariant embedding.
By combining~\eqref{equ:obj:hvae} and~\eqref{equ:obj:asp}, we rewrite the loss as
\begin{equation} \label{equ:obj:dhave}
\begin{aligned}
    \mathcal{L}_{\text{DHVAE}} = & \mathbb{E}_{(\bm{s}, \bm{a}) \sim \mathcal{D}} \bigg[ 
    %\underbrace{
    - \prod_{t=0}^{h} \log \epsilon(a_t|s_t,\proto,\skill)
    %}_{\text{reconstruction loss}} 
    \\
    & + \beta_\rho 
    %\underbrace{
    D_{\text{KL}}(p(\proto) || \protoenc(\proto | \bm{s},\bm{a}))
    %}_{\text{prototype regularization}} 
    &\\
    & + \beta_\sigma 
    %\underbrace{
    D_{\text{KL}}(p(\skill | \proto) || \skillenc(\skill | \proto,\omega))
    %}_{\text{aspect regularization}}) 
    \bigg].
\end{aligned}
\end{equation}
%
% 2. Trainig Priors
For downstream policy learning, we employ a domain-invariant prior $\protopri$ and a domain-variant prior $\skillpri$.
\begin{equation}
    \proto \sim \protopri(s_t), \skill \sim \skillpri(\proto)
\end{equation}
The domain-invariant prior $\protopri$ is conditioned on the current state $s_t$, facilitating the selection of suitable domain-invariant embedding, while the domain-variant prior $\skillpri$ provides a prior distribution over the domain-variant embedding.
By designing a domain-variant prior conditioned solely on the domain-invariant embedding, our model achieves the flexibility to explore a wide range of parameterizations across different domains.
These priors are jointly trained with the hierarchical domain encoder by minimizing the KL divergence between each respective encoder.
\begin{equation} \label{equ:obj:prior}
\begin{aligned}
    \mathcal{L}_{\text{prior}} = & \ \mathbb{E}_{(\bm{s},\bm{a}) \sim \mathcal{D}} [ D_{\text{KL}} \left(\protopri(\proto|s_t) || \protoenc(\proto|\bm{s},\bm{a}) \right) \\
    & + D_{\text{KL}} \left(\skillpri(\skill|\proto) || \skillenc (\skill|\proto,\omega)\right)]
\end{aligned}
\end{equation}
\noindent\textbf{Guided diffusion-based decoder.} To generate diverse skills from domain-invariant embedding $\proto$ and domain-variant embedding $\skill$, we employ the diffusion model~\cite{du:ddpm,durl:imi,durl:off,durl:ddu}.
In particular, we adopt the denoising diffusion probabilistic model (DDPM)~\cite{du:ddpm} to represent our skill decoder.
The decoder reconstructs an action $a_t$ from a noisy input $x^K \sim \mathcal{N}(0,I)$ by sequentially predicting $x^{K-1}, x^{K-2},...,x^{0}(=a_t)$, with each iteration being a marginally denoised version of its predecessor.
During the training phase, the noisy input $x^k$ is generated by progressively adding the Gaussian noise to the original action $a_t(=x^0)$ over $K$ steps as
\begin{equation}
  x^k = \sqrt{\bar{\alpha}^k} a_t + \sqrt{1-\bar{\alpha}^k} \eta
\end{equation}
where $\eta \sim \mathcal{N}(0,I)$ and $\bar{\alpha}^k$ is a variance schedule.
As our skill decoder is designed to predict the noise $\eta$, the reconstruction loss in~\eqref{equ:obj:dhave} becomes an $L_2$ distance between the decoder's output and the noise.
\begin{equation}\label{equ:obj:durec}
    \mathcal{L}_{\text{rec}} = \mathbb{E}_{k \sim [1,K],\eta \sim \mathcal{N}(0,I)} \left[ || \epsilon(x^k,k,s_t,z_\sigma,z_\rho) - \eta ||_2^2 \right]
\end{equation}
To further align with the objective in~\eqref{equ:obj:dhave}, we slightly modify the classifier-free guidance~\cite{du:cfdg} to divide the decoder into two separate parts: domain-invariant decoder $\protodec(x^k,k,s_t,\proto)$ and domain-variant decoder $\skilldec(x^k,k,s_t,\skill)$.
Thus, our decoder is redefined as the combination of both a domain-invariant decoder and a domain-variant decoder, i.e., 
\begin{equation}\label{equ:cfg}
\begin{aligned}
    \epsilon (x^k,k, & s_t,\proto,\skill) := \\
    &  (1-\delta) \epsilon_{\rho}(x^k,k,s_t,z_\sigma) + \delta \epsilon_{\sigma}(x^k,k,s_t,z_\rho)
\end{aligned}
\end{equation}
where $\delta > 0$ serves as a guidance weight that determines the degree of adjustment towards the domain-variant decoder.
This approach allows our domain-invariant decoder to reconstruct actions that consistently execute the designated task across domain features. 
Simultaneously, the domain-variant decoder is seamlessly incorporated by generating guidance that encapsulates domain-variant features.

\begin{algorithm}[t]
    \textbf{Input}: Trainig Datasets $\mathcal{D}$, total denoise step $K$, guidance weight $\delta$, hyperparameters $\beta_\rho$, $\beta_\sigma$
    \begin{algorithmic}[1]
        % Initialization
        \STATE{Initialize encoders $\protoenc$, $\skillenc$, priors $\protopri$, $\skillpri$, decoders $\protodec$, $\skilldec$}
        % Learning
        \WHILE{not converge}
           \STATE{Sample a batch $\{(\bm{s}, \bm{a}, \omega)\}_{i} \sim \mathcal{D}$}
           \STATE{Update $\protoenc$ and $\skillenc$ using $\mathcal{L}_{\text{DHVAE}}$ in~\eqref{equ:obj:dhave}}
            \STATE{Update $\protopri$ and $\skillpri$ using $\mathcal{L}_{\text{prior}}$ in~\eqref{equ:obj:prior}}
           \STATE{Update $\protodec$ and $\skilldec$ using $\mathcal{L}_{\text{rec}}$ in~\eqref{equ:obj:durec}}
        \ENDWHILE
        \STATE{\textbf{return} $\protoenc,\skillenc,\protopri,\skillpri,\protodec,\skilldec$}
    \end{algorithmic}
    \caption{Offline Skill Diffusion} 
    \label{algo:pretrain}
\end{algorithm}
In our framework, we employ the hierarchical domain encoder to establish distinct embedding spaces for domain-invariant and domain-variant features, optimized by the loss defined in~\eqref{equ:obj:dhave}.
Concurrently, the respective priors for the downstream task are jointly trained, optimized by the loss in~\eqref{equ:obj:prior}, as depicted in Figure~\ref{fig:framework} (\romannumeral 1-1), 
Furthermore, to effectively disentangle the domain-invariant and domain-variant embeddings, we jointly train the guided diffusion-based decoder in conjunction with the encoders, optimized by the loss~\eqref{equ:obj:durec}, as illustrated in Figure~\ref{fig:framework} (\romannumeral 1-2).
Algorithm~\ref{algo:pretrain} lists the learning procedures of $\ourmodel$.

\subsection{Downstream Policy Learning}
For efficient policy learning on downstream tasks, we adopt a hierarchical learning scheme akin to other skill-based approaches~\cite{sprl:spirl,sprl:fist}.
In this scheme, the higher-level policy is employed to produce the skill embedding as output rather than directly generating executable actions.
Specifically, we train a policy $\pi(\proto,\skill|s_t)$, designed to align with the hierarchical domain encoder, which generates domain-invariant and domain-variant embeddings, 
\begin{equation}
    \pi(\proto,\skill|s_t) = \pi_{\rho}(\proto|s_t) \cdot \pi_{\sigma}(\skill|\proto).
\end{equation}
Subsequently, these embeddings are fed into the learned guided diffusion-based decoder, which remains frozen to decode them into a sequence of executable actions.
To decode an action from the guided diffusion-based decoder, the process starts with sampling a noisy input $x^K \sim \mathcal{N}(0,I)$. 
Then, the decoder iteratively denoises the input while conditioning the decoder on the disentangled embeddings to generate an action,
\begin{equation}
    x^{k-1} = \frac{1}{\sqrt{\alpha^k}} \left( x^{k} - \frac{1 - \alpha^k}{\sqrt{1-\bar{\alpha}^k}} \epsilon(x^k,k,s_t,\proto,\skill)\right) + \zeta^k \eta
\end{equation}
where $\eta \sim \mathcal{N}(0,I)$, $\zeta^k$ and $\alpha^k$ are parameters for variance schedule.
As it is empirically observed that the low-temperature sampling leads to improved performance, similar to~\cite{durl:ddu}, we set $\zeta^k=0$.

For downstream task adaptation, we explore two different scenarios:  few-shot imitation and online RL adaptation.
For the few-shot imitation, we initialize the high-level policy with the learned domain-invariant and domain-variant priors, and then we fine-tune the high-level policy using \eqref{equ:obj:durec} along with the learned guided diffusion-based decoder.
Likewise, for online RL adaptation, we adopt the soft actor-critic (SAC) algorithm, where the learned priors guide the high-level policy as in~\cite{sprl:spirl}.
In both scenarios, we freeze the decoder and solely fine-tune the high-level policy.
Even in the absence of decoder updates, our framework manages to attain robust performance on downstream tasks in different domains, as demonstrated in Section~\ref{sec:exp:imi}.
Figure~\ref{fig:framework} (\romannumeral 2) illustrates the procedure of downstream policy learning.

\begin{table*}[t]
    \centering
    \small
    \begin{tabular}{lccccccc}
    \toprule
    Domain & Level & BC & SPiRL* & SPiRL-c & SPiRL-c* & FIST* & $\ourmodel$ \\
    \midrule
    \multirow{4}{*}[-6pt]{Speed}
    & Source & $2.66 \pm 0.73$ & $2.16 \pm 0.63$ & $3.62 \pm 0.35$ & $3.72 \pm 0.19$ & $3.98 \pm 0.01$ & $\textbf{3.99} \pm \textbf{0.00}$ \\ \cmidrule{2-8}
    & Level 1 & $2.71 \pm 0.28$ & $1.83 \pm 0.50$ & $3.41 \pm 0.50$ & $3.51 \pm 0.25$ & $3.86 \pm 0.05$  & $\textbf{3.92} \pm \textbf{0.07}$ \\ 
    & Level 2 & $2.62 \pm 0.54$ & $2.16 \pm 0.62$ & $3.33 \pm 0.44$ & $3.24 \pm 0.23$ & $3.51 \pm 0.12$  & $\textbf{3.83} \pm \textbf{0.06}$ \\
    & Level 3 & $2.22 \pm 0.23$ & $2.22 \pm 0.49$ & $2.87 \pm 0.47$ & $3.12 \pm 0.22$ & $3.28 \pm 0.19$ & $\textbf{3.81} \pm \textbf{0.06}$ \\
    \midrule
    \multirow{4}{*}[-6pt]{Energy}
    & Source & $1.49 \pm 0.35$ & $0.67 \pm 0.07$ & $2.83 \pm 0.39$ & $1.60 \pm 0.33$ & $3.08 \pm 0.19$ & $\textbf{3.85} \pm \textbf{0.08}$ \\ \cmidrule{2-8}
    & Level 1 & $1.26 \pm 0.32$ & $0.56 \pm 0.07$ & $1.90 \pm 0.45$ & $1.19 \pm 0.13$ & $2.53 \pm 0.25$ & $\textbf{3.87} \pm \textbf{0.09}$ \\ 
    & Level 2 & $0.90 \pm 0.15$ & $0.47 \pm 0.08$ & $1.77 \pm 0.26$ & $0.91 \pm 0.19$ & $1.85 \pm 0.34$ & $\textbf{3.71} \pm \textbf{0.10}$ \\
    & Level 3 & $0.53 \pm 0.09$ & $1.69 \pm 0.25$ & $0.89 \pm 0.18$ & $2.02 \pm 0.24$ & $1.04 \pm 0.20$ & $\textbf{3.67} \pm \textbf{0.14}$ \\
    \midrule
    \multirow{4}{*}[-6pt]{Wind}
    & Source & $3.32 \pm 0.41$ & $3.78 \pm 0.10$ & $2.83 \pm 0.39$ & $3.92 \pm 0.10$ & $\textbf{4.00} \pm \textbf{0.00}$ & $3.98 \pm 0.02$ \\ \cmidrule{2-8}
    & Level 1 & $2.99 \pm 0.52$ & $3.14 \pm 0.54$ & $1.90 \pm 0.45$ & $3.71 \pm 0.29$ & $\textbf{3.89} \pm \textbf{0.15}$ & $3.78 \pm 0.13$ \\ 
    & Level 2 & $2.19 \pm 0.42$ & $2.62 \pm 0.33$ & $1.77 \pm 0.26$ & $3.24 \pm 0.24$ & $3.24 \pm 0.24$ & $\textbf{3.48} \pm \textbf{0.21}$ \\
    & Level 3 & $2.41 \pm 0.45$ & $2.63 \pm 0.46$ & $0.89 \pm 0.18$ & $3.04 \pm 0.29$ & $3.04 \pm 0.29$ & $\textbf{3.44} \pm \textbf{0.18}$ \\
    \bottomrule
    \end{tabular}
    \caption{Few-shot Imitation Performance in Multi-stage Meta-World: The performance of the baselines and $\ourmodel$ is measured in achieved rewards. For each domain, we categorize domain disparity between the training datasets and downstream tasks into four different levels. The baselines marked with an asterisk (*) indicate that both the high-level policy and decoder are fine-tuned, while the baselines without an asterisk and $\ourmodel$ only fine-tune the high-level policy. Each is evaluated with 3 random seeds, and the highest performance is highlighted in bold.}
    \label{tbl:main:few-shot}
\end{table*}
\section{Evaluations} \label{sec:exp}
\subsection{Experiment Settings}
\noindent\textbf{Environments.}
For evaluation, we use the multi-stage Meta-World, which is implemented based on the Meta-World simulated benchmark~\cite{metaworld}. Each multi-stage task is composed of a sequence of existing Meta-World tasks (sub-tasks).
In these multi-stage tasks, an agent is required to maneuver a robotic arm to complete a series of sub-tasks, such as slide puck, close drawer, and etc. To emulate different domains in the environment, we deliberately add variations to either reward functions or transition dynamics.
Specifically, we modify the reward function by setting time constraints (i.e., speed domains) and by incorporating the energy usage consideration (energy domains).
Furthermore, to emulate varying conditions of transition dynamics in the environment, we manipulate kinematic parameters such as wind (wind domains).

\noindent\textbf{Datasets.} For datasets, we implement several rule-based expert policies tailored to each domain-specific environment. For offline training datasets, we collect 20 trajectories for each source domain (of $6 \sim 16$ domains). For few-shot imitation datasets, we collect another 3 trajectories for each target domain.

\noindent\textbf{Baselines.} For baselines, we implement several imitation learning and skill-based RL algorithms.
\begin{itemize}
    \item BC~\cite{bc} is a widely used supervised behavior cloning method. A policy is learned on the training datasets and then fine-tuned for the downstream tasks.
    \item SPiRL~\cite{sprl:spirl} is a state-of-art skill-based RL algorithm that employs a hierarchical structure to embeds skills into a latent space, thereby accelerating the downstream adaptation. 
    \item SPiRL-c is a variant of SPiRL that uses the closed-loop skill decoder, used in~\cite{sprl:skild}. 
    \item FIST~\cite{sprl:fist} is a state-of-art few-shot skill-based imitation algorithm that employs a semi-parametric approach for skill determination.
\end{itemize}
To obtain few-shot imitation, the baselines such as BC, SPiRL and SPiRL-c are pretrained on the training datasets and then fine-tuned on the few-shot imitation datasets.

\subsection{Few-shot Imitation Performance}\label{sec:exp:imi}
Table~\ref{tbl:main:few-shot} compares the few-shot imitation performance in rewards achieved by our framework ($\ourmodel$) and other baselines (BC, SPiRL, SPiRL-c, FIST).
Among the baselines, those denoted with an asterisk (*) signify that both the high-level policy and the decoder are fine-tuned, while those without an asterisk indicate fine-tuning solely of the high-level policy.
An important distinction is that $\ourmodel$ exclusively focuses on fine-tuning the high-level policy.
Based on the domain disparities between the datasets and the downstream tasks, we categorize them into four different levels.
At the source level, downstream tasks remain same with the source domains. As we move to higher levels, the domain disparities become more pronounced.

As shown, the baselines exhibit a notable decline in performance as the domain dissimilarity increases from source to level 4, where even the most competitive baseline FIST achieves an average degradation of $25.3\%$. 
In contrast, $\ourmodel$ consistently maintains robust performance across all domains and levels, achieving a small degradation of $7.6\%$ at average.

In this experiment, SPiRL-c demonstrates relatively low performance, primarily because its decoder can only generate skills present in its training data. Consequently, this poses a challenge in attaining robust performance when solely relying on fine-tuning the high-level policy.
Meanwhile, SPiRL-c* is expected to yield higher performance than SPiRL-c as it fine-tunes both the high-level policy and the decoder; yet, in some cases, SPiRL-c surpasses SPiRL-c*.
This is because fine-tuning the entire model with a few samples might cause a covariant shift, a phenomenon commonly observed in the few-shot imitation studies~\cite{sprl:fist,sprl:sailor}.
FIST* adopts a different strategy, involving the fine-tuning of the entire model along with the utilization of a semi-parametric method to retrieve the future state $s_{t+H}$ that it aims to reach from the training datasets.
While the semi-parametric method leads to improved performance for tasks in source domains, FIST* is prone to fail in producing skills for downstream tasks in different domains. This is because the training datasets do not cover the skills required for different domains.
In contrast, $\ourmodel$ disentangles the domain-invariant and -variant features to effectively generate the skills through the guided diffusion-based decoder. This allows for robust performance in few-shot adaptation across different domains, through fine-tuning solely the high-level policy.

\subsection{Analysis}
\noindent\textbf{Online RL.}
Table~\ref{tbl:main:rl} compares online RL adaptation performance in reward achieved by our framework ($\ourmodel$) and other baselines (BC+SAC, SPiRL, SPiRL-c).
Both the baselines and $\ourmodel$ fine-tune the high-level policy via  the SAC algorithm.
Here, we categorize domain disparities into source and target domains, where target domains correspond to the level 3 in Table~\ref{tbl:main:few-shot}.
For more stable learning in target domains, we warm-up the high-level policy with a single trajectory for $\ourmodel$ and other baselines.
The performance gap between SPiRL-c and $\ourmodel$ is not remarkable in source domain tasks, as expected.
In contrary, $\ourmodel$ exhibits superior performance compared to the baselines for tasks in different domains, outperforming SPiRL-c by $89.16\%$.
This highlights the capability of our guided diffusion-based decoder in generating diverse skills that extend beyond the limitations of the given datasets.
\begin{table}[h]
    \centering
    \small
    \begin{tabular}{lccc}
    \toprule
    Level & BC+SAC & SPiRL-c & $\ourmodel$ \\
    \midrule
    Source & $0.00 \pm 0.00$ & $4.00 \pm 0.00$ & $\textbf{4.00} \pm \textbf{0.00}$ \\
    Target & $0.00 \pm 0.00$ & $0.36 \pm 0.02$ & $\textbf{3.32} \pm \textbf{0.20}$ \\
    \bottomrule
    \end{tabular}
    \caption{Online RL Performance in Speed Domain}
    \label{tbl:main:rl}
\end{table}

\begin{figure}[ht]
    \centering
    \includegraphics[width=.9\linewidth]{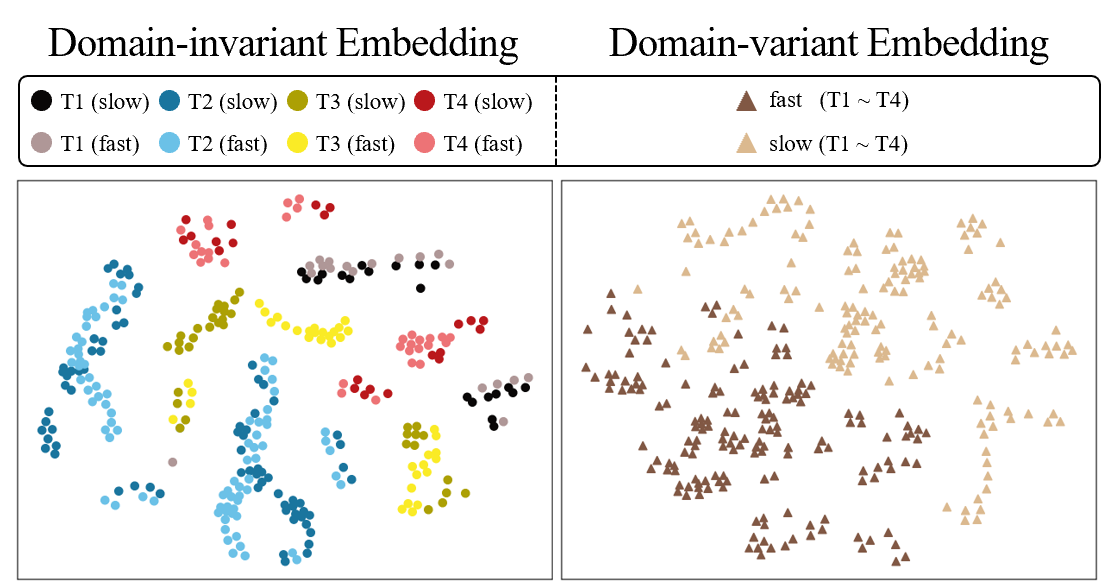}
    \caption{Visualization of Domain-invariant and Domain-variant embeddings}
    \label{fig:ana:emb}
\end{figure}

\noindent\textbf{Embedding Visualization.} Here, we verify the efficacy of our $\ourmodel$ framework in disentangled embeddings. 
Figure~\ref{fig:ana:emb} visualizes the domain-invariant and domain-variant embeddings generated by $\ourmodel$ separately in two distinct speed domains (fast and slow).
In the figure, the labels T1 to T4 correspond to sub-tasks numbered from 1 to 4. Regarding the domain-invariant embeddings, we observe that the identical tasks are paired together, thereby establishing the domain-invariant knowledge. On the other hand, the domain-variant embeddings are grouped with respect to the domains, implying the encapsulation of domain-variant knowledge.
This indicates the effectiveness of our offline skill diffusion process, which disentangles domain-invariant and variant features.

\begin{figure}[ht]
    \centering
    \subfigure[Few-shot imitation]{
        \centering
        \includegraphics[width=0.47\linewidth]{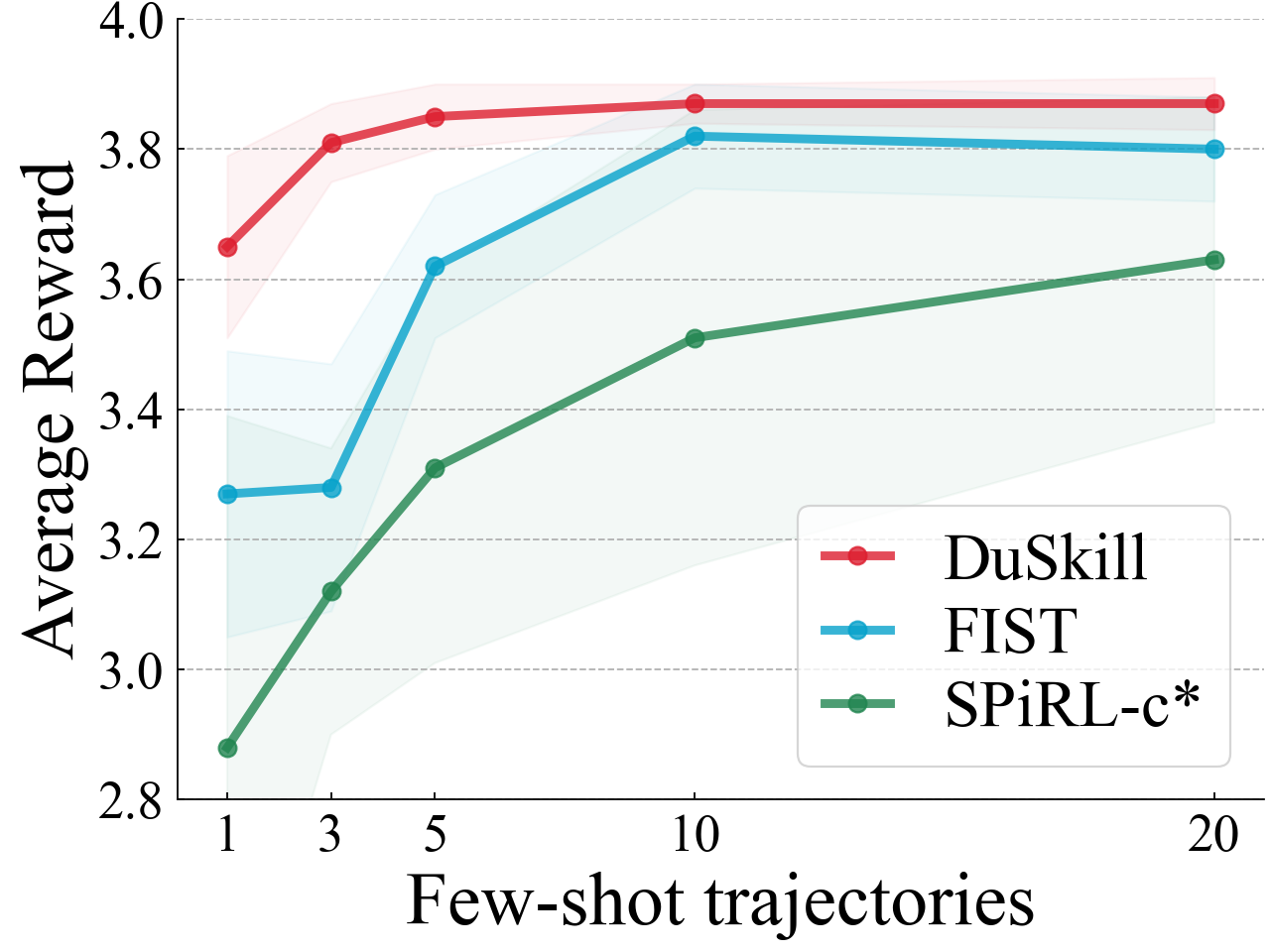}
        \label{fig:ana:eff:imi}
    }
    \subfigure[Online RL]{
        \centering
        \includegraphics[width=0.47\linewidth]{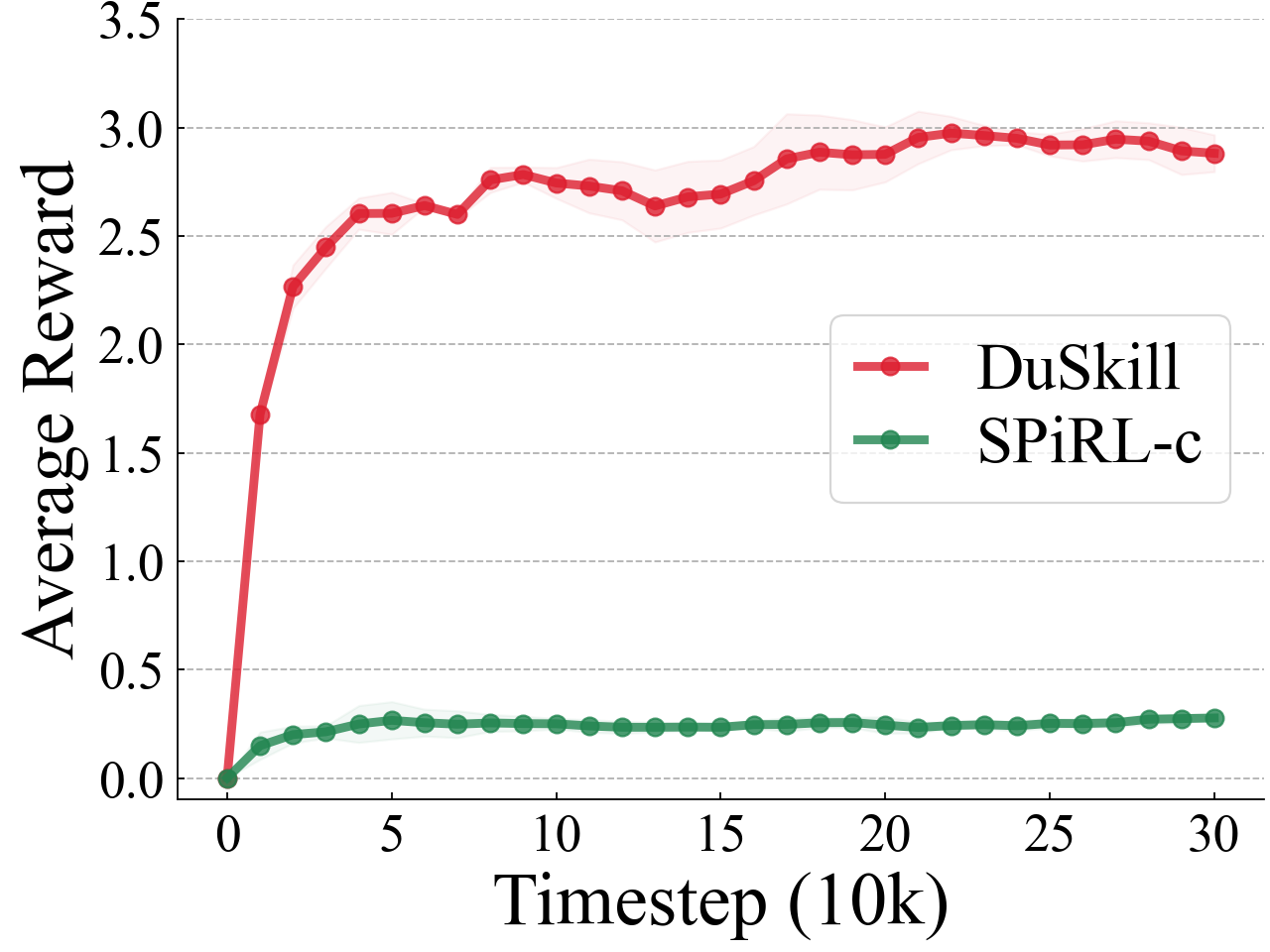}
        \label{fig:ana:eff:onl}
    }
    \caption{Sample Efficiency of Downstream Policy Learning}
    \label{fig:ana:eff}
\end{figure}

\noindent\textbf{Sample Efficiency.} Figure~\ref{fig:ana:eff} shows the performance with respect to samples (or timesteps) used by $\ourmodel$ and the baselines (SPiRL-c, SPiRL-c*, FIST) for downstream policy learning in few-shot imitation and online RL scenarios.
For the few-shot imitation learning, we test with different numbers of few-shot trajectories $(1\sim20)$.
As shown in Figure~\ref{fig:ana:eff:imi}, $\ourmodel$ exhibits robust performance with only a $4.89\%$ drop from $1$ to $20$ trajectories, whereas the most competitive baseline FIST shows a notable performance drop of $13.96\%$.
Furthermore, as shown in Figure~\ref{fig:ana:eff:onl}, $\ourmodel$ efficiently adapts to the downstream task in online settings, enhancing performance with only $50$k samples, while SPiRL-c rarely achieves improvement with those samples.

\noindent\textbf{Ablation on Skill Diffusion.} 
Table~\ref{tbl:abl:1} provides an ablation study of $\ourmodel$, focusing on the impact of the hierarchical embedding structure and guided diffusion-based decoder. In this study, we conduct few-shot imitation scenarios with two ablated variants. DU is a variant of SPiRL-c, utilizing a naive diffusion model for the skill decoder, while HDU employs the hierarchical embedding structure like $\ourmodel$ and a naive diffusion model for the skill decoder.
The results show that the combination of hierarchical domain encoder and guided diffusion based decoder in $\ourmodel$ yields improved performance compared to the other variants, showing $8.77 \sim 8.92\%$ average performance gains.
This is not surprising since the guided diffusion based decoder promotes the learning of disentangled representations, as the domain-invariant and domain-variant decoders are conditioned on each embedding to generate executable actions effectively.
Therefore, it is crucial to employ both hierarchical embedding structure and guided diffusion based decoder for achieving better few-shot imitation learning performance.
\begin{table}[h]
    \centering
    \small
    \begin{tabular}{cccc}
    \toprule
    \ Domain \ & DU & HDU & $\ourmodel$ \\
    \midrule
    Speed   & $3.59 \pm 0.19$ & $3.49 \pm 0.68$ & $\textbf{3.81} \pm \textbf{0.06}$ \\
    Energy  & $3.16 \pm 0.35$ & $3.39 \pm 0.36$ & $\textbf{3.67} \pm \textbf{0.14}$ \\
    Wind    & $3.19 \pm 0.17$ & $3.08 \pm 0.36$ & $\textbf{3.44} \pm \textbf{0.18}$ \\
    \bottomrule
    \end{tabular}
    \caption{Performance by Encoder and Decoder Types}
    \label{tbl:abl:1}
\end{table}

\section{Related Work}
\noindent\textbf{Skill-based Learning.} To leverage offline datasets for long-horizon tasks, hierarchical skill representation learning techniques have been investigated in the context of online RL~\cite{sprl:spirl,sprl:skild} and imitation learning~\cite{sprl:fist,sprl:sailor,sprl:ret}.
\citet{sprl:spirl} proposed the hierarchical skill learning structure to accelerate the downstream task adaptation by guiding a high-level policy with learned skill priors.
Meanwhile, \citet{sprl:fist} exploited a semi-parametric approach within this hierarchical skill structure, focusing on the few-shot imitation.
In our work, we also utilize skill embedding techniques, but we tackle the challenge of adapting to downstream tasks in different domains.
Unlike the prior work, our $\ourmodel$ adapts a robust generative model such as diffusion with the technique of disentangled skill embeddings, enabling the effective generation of diverse skills in offline.

\noindent \textbf{Diffusion for RL.} Given the remarkable success of diffusion models in the field of computer vision~\cite{du:ddpm,du:stabledu}, their application has been extended to sequential decision-making problems in recent years.
\citet{durl:imi} utilized diffusion models to imitate human demonstrations, capitalizing on their capacity to represent highly multi-modal data.
\citet{durl:off} adopted diffusion models in the context of offline RL to implement policy regularization. 
Furthermore, \citet{durl:adapt} leveraged diffusion models to generate diverse synthetic trajectories on limited training data, aiming to have  self-evolving offline RL for goal-conditioned RL tasks.
Recently, \citet{durl:ddu} proposed a general framework for sequential decision-making problems using diffusion models. It allows for dynamic recombination of behaviors at test time by conditioning the diffusion models on several factors such as returns, constraints, or skills.
%
% Yet, this framework has limited capabilities for skill learning in terms of diversity in that it only accommodates a predefined set of skills.
%
To the best of our knowledge, our $\ourmodel$ framework is the first to integrate a diffusion model with skill embedding techniques, providing a novel hierarchical RL method to generate diverse skills on limited datasets and achieving robust performance for downstream tasks in different domains.

\section{Conclusion}
In this work, we presented the $\ourmodel$ framework, a novel approach designed to bridge the gap between the given datasets and downstream tasks that exist within different domains.
We devised the offline skill diffusion, which employs the guided diffusion-based decoder in conjunction with the hierarchical encoders to effectively disentangles domain-invariant and -variant features from skills. This enables the generation of diverse skills capable of addressing tasks in different domains.
Our framework stands apart from existing skill-based learning approaches, which are typically limited to adapt within the domains encountered during skill pretraining.
In our future work, we aim to extend our framework to address challenging cross-domain situations with significant domain shifts, such as entirely different tasks, robot embodiment variations, or different simulation environments. 
%

% \appendix

\section*{Acknowledgments}
We would like to thank anonymous reviewers for theri valuable comments and suggestions.
This work was supported by Institute of Information \& communications Technology Planning \& Evaluation (IITP) grant funded by the Korea government (MSIT) (No. 2022-0-01045, 2022-0-00043, 2019-0-00421, 2021-0-00875) and by the National Research Foundation of Korea (NRF) grant funded by the MSIT (No. RS-2023-00213118).

\bibliography{aaai24}

\newpage 
\appendix

\appendix

\section{Benchmark Environments}
\subsection{Multi-stage Meta-World}
\begin{figure}[h]
    \centering
    \includegraphics[width=.7\linewidth]{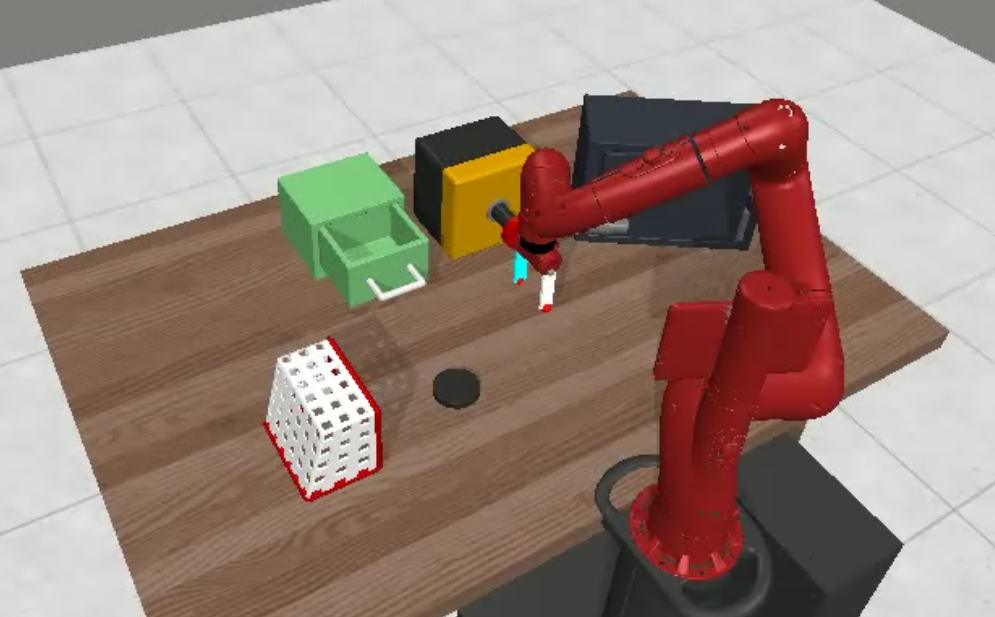}
    \caption{
    Multi-stage Meta-World with $4$ sub-tasks
    }
    \label{fig:app:metaworld}
\end{figure}

We modify the Meta-World environment to incorporate multi-stage long-horizon tasks, which we refer to as Multi-stage Meta-World. In this modified environment, an agent is required to  execute a series of sub-tasks sequentially.
These sub-tasks are comprised of existing Meta-World sub-tasks, such as sliding a plate, closing a drawer, pushing a button, and opening a door. In our implementation, we assemble four such sub-tasks to create a single, composite task.
The dimensionality of state space is $\StateSet \in\mathbb{R}^{140}$ consisting of the position of the robot arm and the objects, and the dimensionality of action space is $\ActionSet \in \mathbb{R}^{4}$ consisting of directional vector of size $3$ applied to end-effector and torque vector of size $1$ applied to the gripper.
To generate expert datasets, we implement a heuristic algorithm for each domain.
Figure~\ref{fig:app:metaworld} depicts the example image of Multi-stage Meta-World.

\noindent\textbf{Speed domain.}
In the speed domain, the agent maneuvers the velocity of the robot arm. 
In this setting, the agent is required to complete each sub-task within a given timesteps.

\noindent\textbf{Energy domain.}
In the energy domain, the agent manipulates the acceleration applied to the robot arm.
In this setting, the agent is required to complete each sub-task within a given energy capacity.

\noindent\textbf{Wind domain.}
In the wind domain, the strength of the wind varies every fixed timesteps.
In this setting, the agent is required to complete each sub-task successfully.

\subsection{Expert Datasets}
To generate expert datasets, we implement heuristic algorithms for each domain.
Here, we illustrate the dataset settings used for varying levels of domain disparity with the sub-task sequence and domain parameterization($\omega$) associated with each sub-tasks.
Note that expert datasets from source domains are all used for pretraining the skill representation, while few-shot datasets are used individually for imitating each task.

\noindent\textbf{Source-level.}
Expert datasets from source domains are:
\begin{itemize}
    \item puck-door-button-drawer : 4-4-4-4 (domain parameter)
    \item puck-drawer-door-button : 4-4-4-4
    \item drawer-puck-door-button : 4-4-4-4
    \item drawer-door-puck-button : 4-4-4-4
    \item button-puck-door-drawer : 4-4-4-4
    \item button-door-puck-drawer : 4-4-4-4
\end{itemize}
Few-shot datasets from target domains are:
\begin{itemize}
    \item puck-button-drawer-door : 4-4-4-4
    \item drawer-puck-button-door : 4-4-4-4
    \item button-drawer-puck-door : 4-4-4-4
\end{itemize}

\noindent\textbf{Level-1.}
Expert datasets from source domains are:
\begin{itemize}
    \item puck-door-button-drawer : 2-6-2-6
    \item puck-drawer-door-button : 6-6-2-2
    \item drawer-puck-door-button : 6-2-2-2
    \item drawer-door-puck-button : 6-6-6-2
    \item button-puck-door-drawer : 2-6-2-6
    \item button-door-puck-drawer : 2-6-2-6
\end{itemize}
Few-shot datasets from target domains are:
\begin{itemize}
    \item puck-button-drawer-door : 2-6-6-2
    \item drawer-puck-button-door : 2-2-2-2
    \item button-drawer-puck-door : 6-6-6-6
\end{itemize}

\noindent\textbf{Level-2.}
Expert datasets from source domains are:
\begin{itemize}
    \item puck-door-button-drawer : 2-0-6-6, 4-4-2-0
    \item puck-drawer-door-button : 4-2-0-0, 6-8-8-6
    \item drawer-puck-door-button : 2-2-0-0, 8-4-8-6
    \item drawer-door-puck-button : 6-0-2-0, 2-4-6-2
    \item button-puck-door-drawer : 6-6-8-6. 0-4-4-2
    \item button-door-puck-drawer : 2-8-6-8, 6-0-4-6
\end{itemize}
Few-shot datasets from target domains are:
\begin{itemize}
    \item puck-button-drawer-door : 0-8-0-2, 8-8-4-6, 8-4-4-2
    \item drawer-puck-button-door : 0-8-4-2, 4-8-4-6. 4-0-6-8
    \item button-drawer-puck-door : 4-0-8-2, 8-4-8-6, 8-0-8-4
\end{itemize}

\noindent\textbf{Level-3.}
Expert datasets from source domains are:
\begin{itemize}
    \item puck-door-button-drawer : 2-0-2-6, 6-4-0-8
    \item puck-drawer-door-button : 6-6-0-2, 2-8-8-0
    \item drawer-puck-door-button : 6-2-8-0, 8-6-0-2
    \item drawer-door-puck-button : 8-0-2-0, 6-4-6-2
    \item button-puck-door-drawer : 2-6-8-6. 0-2-4-8
    \item button-door-puck-drawer : 2-8-6-8, 0-0-2-6
\end{itemize}
Few-shot datasets from target domains are:
\begin{itemize}
    \item puck-button-drawer-door : 0-6-2-2, 4-8-0-6, 8-4-6-6
    \item drawer-puck-button-door : 2-0-6-6, 4-8-8-2. 0-4-4-6
    \item button-drawer-puck-door : 4-2-0-2, 6-0-8-6, 8-4-4-2
\end{itemize}

\section{Implementation Details}
In this section, we describe our expert datasets generation procedure, and show implementation details of the baselines and $\ourmodel$ with hyperparameter settings used for training. 
We use the open source projects JAX with Haiku throughout our implementation, and for experiments, we use a system of an NVIDIA RTX A6000 GPU and an Intel(R) Core(TM) i9-10980XE CPU.

\subsection{BC}
We implement BC using the supervised behavior cloning algorithm~\cite{bc}.
We pretrain BC model with the expert datasets from source domains, and fine-tune it with few-shot datasets.
The hyperparameter settings for BC are summarized in Table~\ref{tbl:bc_params}.
\begin{table}[h]
    \small
    \centering
    \begin{tabular}{lc}
    \toprule
    \textbf{HyperParameter} & \textbf{Value} \\
    \midrule
    Learning rate       & $1 \times 10^{-4}$ \\
    Batch size          & $128$ \\
    Timestep            & $1 \times 10^5$  \\
    Actor network       & $5$ layers of $128$ size MLP \\
    \bottomrule
    \end{tabular}
    \caption{Hyperparameter settings for BC}
    \label{tbl:bc_params}
\end{table}

For online RL, we employ SAC algorithm pretrained with BC (BC+SAC), where the actor network of SAC is initialized with pretrained BC model. The hyperparameter settings for BC+SAC are summarized in Table~\ref{tbl:bc+sac_params}.
\begin{table}[h]
    \small
    \centering
    \begin{tabular}{lc}
    \toprule
    \textbf{HyperParameter} & \textbf{Value} \\
    \midrule
    Learning rate       & $3 \times 10^{-4}$ \\
    Batch size          & $64$ \\
    Timestep            & $3 \times 10^5$ \\
    Actor network       & $5$ layers of $128$ size MLP \\
    Critic network       & $5$ layers of $128$ size MLP \\
    \bottomrule
    \end{tabular}
    \caption{Hyperparameter settings for BC+SAC}
    \label{tbl:bc+sac_params}
\end{table}

\subsection{SPiRL}
We implement SPiRL which consists of an encoder, a prior, and a decoder.
Depending on the type of decoder, we also implement SPiRL-c which utilizes a closed-loop decoder conditioned on current state~\cite{sprl:skild}.
We pretrain the whole model with the expert datasets from source domains, and train the high-level policy initialized with the prior on few-shot datasets.
For SPiRL* and SPiRL-c*, we further fine-tune the decoder.
The hyperparameter settings for SPiRL are summarized in Table~\ref{tbl:spirl_params}. 
\begin{table}[h]
    \small
    \centering
    \begin{tabular}{lc}
    \toprule
    \textbf{HyperParameter} & \textbf{Value} \\
    \midrule
    Learning rate       & $1 \times 10^{-3}$ \\
    Batch size          & $128$ \\
    Timestep            & $1 \times 10^5$ \\
    Encoder network     & $5$ layers of $128$ size MLP \\
    Prior network       & $5$ layers of $128$ size MLP \\
    Decoder network     & $5$ layers of $128$ size MLP \\
    skill length ($h$)  & $10$ \\
    latent size         & $32$ \\
    regularization term ($\beta$) & $5 \times 10^{-4}$ \\
    \bottomrule
    \end{tabular}
    \caption{Hyperparameter settings for SPiRL}
    \label{tbl:spirl_params}
\end{table}

For online RL, we employ prior regularized SAC~\cite{sprl:spirl}, where a high-level policy is learned through regularizing with the learned prior. The hyperparameter settings for online RL in SPiRL are identical to Table~\ref{tbl:bc+sac_params}.

\subsection{FIST}
We implement FIST which consists of an encoder, a prior, a distance model, and a decoder.
Unlike SPiRL, the prior in FIST is an inverse skill dynamics model that takes both $s_t$ and $s_{t+h}$ as inputs to generate a skill, and a distance function is trained with contrastive loss~\cite{infonce}.
We pretrain the entire model with the expert datasets from source domains, and train the high-level policy initialized with the prior and the decoder on few-shot datasets. 
During inference, $s_{t+h}$ is retrieved from the source datasets and then fed into the high-level policy with the current state for decision making.
The hyperparameter settings for FIST are summarized in Table~\ref{tbl:fist_params}.
\begin{table}[h]
    \small
    \centering
    \begin{tabular}{lc}
    \toprule
    \textbf{HyperParameter} & \textbf{Value} \\
    \midrule
    Learning rate       & $1 \times 10^{-3}$ \\
    Batch size          & $128$ \\
    Timestep            & $1 \times 10^5$ \\
    Encoder network     & $5$ layers of $128$ size MLP \\
    Prior network       & $5$ layers of $128$ size MLP \\
    Decoder network     & $5$ layers of $128$ size MLP \\
    skill length ($h$)  & $10$ \\
    latent size         & $32$ \\
    regularization term ($\beta$) & $5 \times 10^{-4}$ \\
    \bottomrule
    \end{tabular}
    \caption{Hyperparameter settings for SPiRL}
    \label{tbl:fist_params}
\end{table}

\subsection{$\ourmodel$}
For $\ourmodel$, we implement a hierarchical domain encoder, a pair of priors, and a diffusion-based decoder.
Unlike SPiRL, $\ourmodel$ utilizes a guided diffusion model for skill decoder.
We pretrain the entire model with the expert datasets from source domains, and train the high-level policy initialized with the priors on few-shot datasets.
The hyperparameter settings for FIST are summarized in Table~\ref{tbl:duskill_params}.
\begin{table}[h]
    \small
    \centering
    \begin{tabular}{lc}
    \toprule
    \textbf{HyperParameter} & \textbf{Value} \\
    \midrule
    Learning rate       & $1 \times 10^{-3}$ \\
    Batch size          & $128$ \\
    Timestep            & $1 \times 10^5$ \\
    Encoder network     & $5$ layers of $128$ size MLP \\
    Prior network       & $5$ layers of $128$ size MLP \\
    Decoder network     & $5$ layers of $128$ size MLP \\
    skill length ($h$)  & $10$ \\
    latent size         & $32$ \\
    $\beta_\rho$        & $5 \times 10^{-4}$ \\
    $\beta_\sigma$      & $1 \times 10^{-4}$ \\
    Denoising timesteps ($K$) & $50$ \\
    Guidance weight ($\delta$) & $0.5$ \\
    Variance scheduler  & linear scheduler \\
    \bottomrule
    \end{tabular}
    \caption{Hyperparameter settings for SPiRL}
    \label{tbl:duskill_params}
\end{table}

For online RL, we employ the same method as in SPiRL, and the hyperparameter settings are identical to Table~\ref{tbl:bc+sac_params}.

\end{document}